\documentclass[11pt,a4paper]{article}
\usepackage[hyperref]{emnlp2020}
\usepackage{times}
\usepackage{latexsym}
\usepackage{xspace}
\usepackage{amsmath}
\usepackage{graphicx}
\usepackage{subcaption}
\usepackage{multirow}
\usepackage{pifont}

\usepackage{microtype}

\aclfinalcopy 

\newcommand{\method}{PALM\xspace}
\newcommand{\xmark}{\ding{55}}
\title{\method: Pre-training an Autoencoding\&Autoregressive Language Model for Context-conditioned Generation}

\author{Bin Bi, Chenliang Li, Chen Wu, Ming Yan, \\
\textbf{Wei Wang, Songfang Huang, Fei Huang, Luo Si} \\
  Alibaba Group \\
  {\tt \{b.bi, lcl193798, wuchen.wc, ym119608\}@alibaba-inc.com} \\
  {\tt \{hebian.ww, songfang.hsf, f.huang, luo.si\}@alibaba-inc.com}}

\date{}

\begin{document}
\maketitle
\begin{abstract}
Self-supervised pre-training, such as BERT~\cite{bert2018jacob}, MASS~\cite{mass2019song} and BART~\cite{bart2019}, has emerged as a powerful technique for natural language understanding and generation. Existing pre-training techniques employ autoencoding and/or autoregressive objectives to train Transformer-based models by recovering original word tokens from corrupted text with some masked tokens. The training goals of existing techniques are often inconsistent with the goals of many language generation tasks, such as generative question answering and conversational response generation, for producing new text given context.

This work presents \method with a novel scheme that jointly pre-trains an autoencoding and autoregressive language model on a large unlabeled corpus, specifically designed for generating new text conditioned on context. The new scheme alleviates the mismatch introduced by the existing denoising scheme between pre-training and fine-tuning where generation is more than reconstructing original text. An extensive set of experiments show that \method achieves new state-of-the-art results on a variety of language generation benchmarks covering generative question answering (Rank 1 on the official MARCO leaderboard), abstractive summarization on CNN/DailyMail as well as Gigaword, question generation on SQuAD, and conversational response generation on Cornell Movie Dialogues.

\end{abstract}

\section{Introduction}
Self-supervised pre-training has achieved great success in a wide range of natural language understanding (NLU) tasks~\cite{semi2015dai,howard-ruder-2018-universal,Radford2018ImprovingLU,peters-etal-2018-deep,bert2018jacob}. Different from language understanding, language generation aims at generating natural language sentences, including tasks like neural machine translation~\cite{Bahdanau:2015,Vaswani17attention}, abstractive summarization~\cite{rush-etal-2015-neural,See:17,Gehrmann:18}, generative question answering (QA)~\cite{Tan:17,bi2019incorporating}, question generation~\cite{zhao-etal-2018-paragraph} and conversational response generation~\cite{Vinyals:2015}. Many of the language generation tasks require the models to read and to comprehend a given document, based on which output text is generated. In this paper, we present \method, a novel approach to \textbf{P}re-training an \textbf{A}utoencoding\&autoregressive \textbf{L}anguage \textbf{M}odel for text generation based on reading comprehension of textual context.

Recently, several pre-training methods have been proposed for language generation. GPT~\cite{Radford2018ImprovingLU} and GPT-2~\cite{gpt2} use a left-to-right Transformer decoder to generate a text sequence token-by-token, which lacks an encoder to condition generation on context. In contrast, MASS~\cite{mass2019song} and BART~\cite{bart2019} both employ a Transformer-based encoder-decoder framework, with a bidirectional encoder over corrupted (masked) text and a left-to-right decoder reconstructing the original text. While such denoising pre-training objectives work well for the downstream generation tasks where generated text comes from input but is manipulated, they are less related to the comprehension-based generation tasks asking for instead generating continuations, responses or answers by comprehending input context.

\method is specifically designed to pre-train a backbone model on a large unlabeled corpus for fine-tuning on the downstream comprehension-based generation tasks, one example of which is generative QA. In generative question answering, QA models are asked to generate an abstractive answer in natural language to a given question by reading and comprehending a contextual passage. Abstractive answer generation is more than manipulating tokens in the passage. An abstractive answer reflects the understanding of the passage and the question, and can include content out of the passage to be self-contained and well-formed. To address comprehension-based generation like generative QA, \method uses the pre-training objectives that are closely related to the downstream tasks. Specifically, it differs from existing generative pre-training methods in that \method goes beyond the solely autoencoding/autoregressive methods and combines the merits of autoencoding and autoregression in a single framework. Moreover, it possesses a mechanism built in pre-training for generating coherent text from given context.

With the new design, \method surpasses existing language generation methods with or without pre-training -- It was trained on 16 NVIDIA V100 GPUs for 3 days in our experiments, and expected to perform even better if trained for longer. \method gives surprisingly good empirical results on a variety of context-aware generation tasks, including pushing the state-of-the-art Rouge-L on the \emph{MARCO Natural Language Generation} benchmark to 0.498 (Rank 1 on the leaderboard~\footnote{\url{http://www.msmarco.org/leaders.aspx}}) and on Gigaword summarization to 36.75, as well as establishing the state-of-the-art ROUGE-1 (44.30) and ROUGE-L (41.41) on CNN/Daily Mail.

We make the following major contributions in this paper:
\begin{itemize}
    \item We propose \method, a novel approach to pre-training a language model on a large unlabeled text corpus, which is able to comprehend contextual text. The pre-trained model is particularly effective to be fine-tuned for language generation conditioned on context.
    \item \method significantly advances the state-of-the-art results on a variety of language generation applications, including generative QA, abstractive summarization, question generation, and conversational response generation. It clearly demonstrates \method's effectiveness and generalizability in language generation.
\end{itemize}

\section{\method for Context-conditioned Generation}
This section presents the new mechanism and pre-training objectives of \method for generation conditioned on context. The differences between \method and prior pre-training approaches are discussed as well.

\subsection{Joint Modeling of Autoencoding and Autoregression}
We denote $(x,y)\in (\mathcal{X},\mathcal{Y})$ as a pair of text pieces, where $x=(x_1,x_2,\dots,x_m)$ is the source text with $m$ tokens, and $y=(y_1,y_2,\dots,y_n)$ is the target text with $n$ tokens. $\mathcal{X}$ and $\mathcal{Y}$ denote the sets of source text and target text, respectively. \method uses the standard Transformer encoder-decoder from~\cite{Vaswani17attention} as the base architecture, which maximizes the log-likelihood objective: $\mathcal{L}(\theta;(\mathcal{X},\mathcal{Y}))=\sum_{(x,y)\in (\mathcal{X},\mathcal{Y})}\log P(y|x;\theta)$.

Existing Transformer-based pre-training methods employ either autoencoding or autoregressive objectives for self-supervision. Autoencoding-based pre-training aims to reconstruct the original text from corrupted input. Notable examples are BERT and its variants RoBERTa and ALBERT, where a certain portion of input tokens are replaced by a special symbol \texttt{[MASK]}. The models are trained to recover the original tokens from the corrupted version by utilizing bidirectional context. However, these autoencoding methods are not applicable to text generation where bidirectional contexts are not available.

On the other hand, an autoregressive model, such as GPT~\cite{Radford2018ImprovingLU,gpt2}, is only trained to encode unidirectional context (either forward or backward). Specifically, at each output timestep, a token is sampled from the model’s predicted distribution and the sample is fed back into the model to produce a prediction for the next output timestep, and so on. While applicable to text generation, the autoregressive methods are not effective at modeling deep bidirectional context. On the contrary, downstream generation tasks often ask a model to condition generation on given textual context. This results in a gap between autoregressive modeling and effective pre-training.

\begin{figure*}[t]
     \centering
     \begin{subfigure}{0.47\textwidth}
         \centering
         \includegraphics[width=\textwidth]{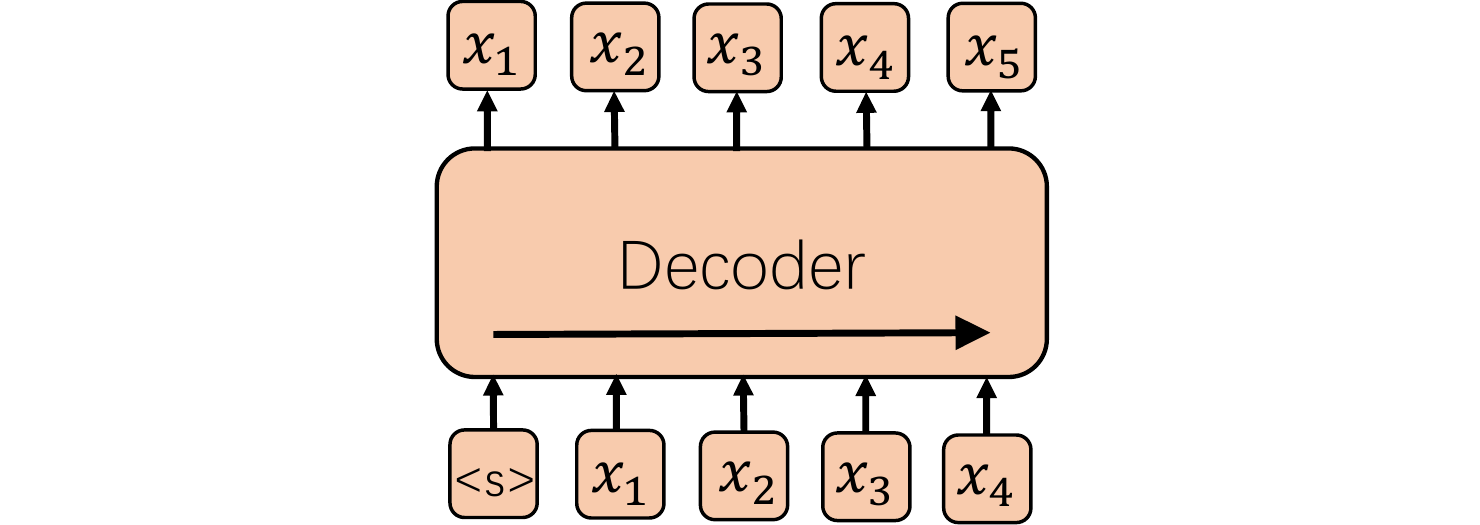}
         \caption{GPT: Tokens are predicted autoregressively, meaning that GPT can be used for generation. However, it lacks an encoder to condition generation on context.}
         \label{fig:gpt}
     \end{subfigure}
     \hfill
     \begin{subfigure}{0.47\textwidth}
         \centering
         \includegraphics[width=\textwidth]{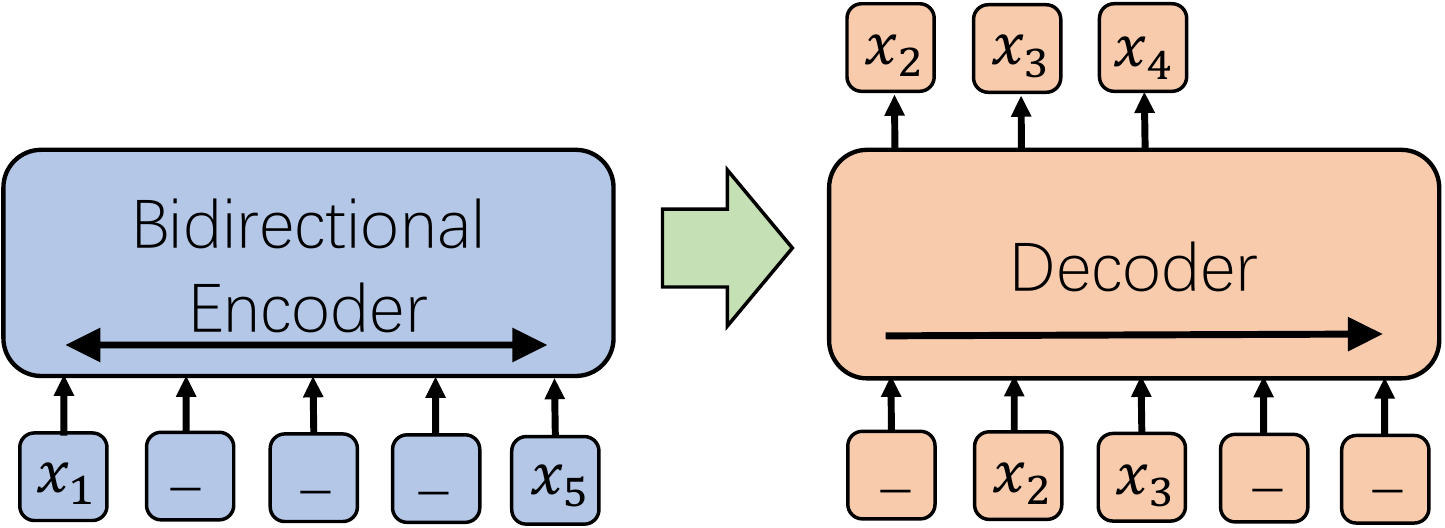}
         \caption{MASS: It is based on the encoder-decoder architecture, but the decoder predicts only the tokens that are masked out in the text input to the encoder.}
         \label{fig:mass}
     \end{subfigure}
     \bigskip\\
     \begin{subfigure}{0.47\textwidth}
         \centering
         \includegraphics[width=\textwidth]{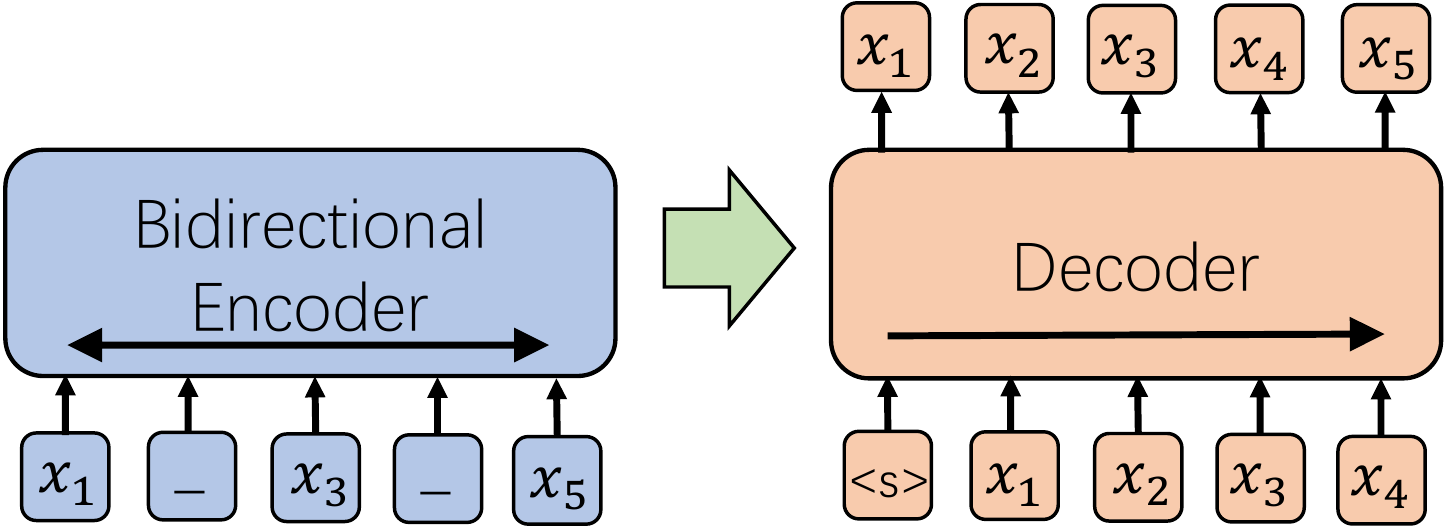}
         \caption{BART: Rather than masked tokens, the decoder reconstructs the original full sentence from the corrupted input to the encoder. However, it mismatches with most downstream generation which is more than reconstructing original input.}
         \label{fig:bart}
     \end{subfigure}
     \hfill
     \begin{subfigure}{0.47\textwidth}
         \centering
         \includegraphics[width=\textwidth]{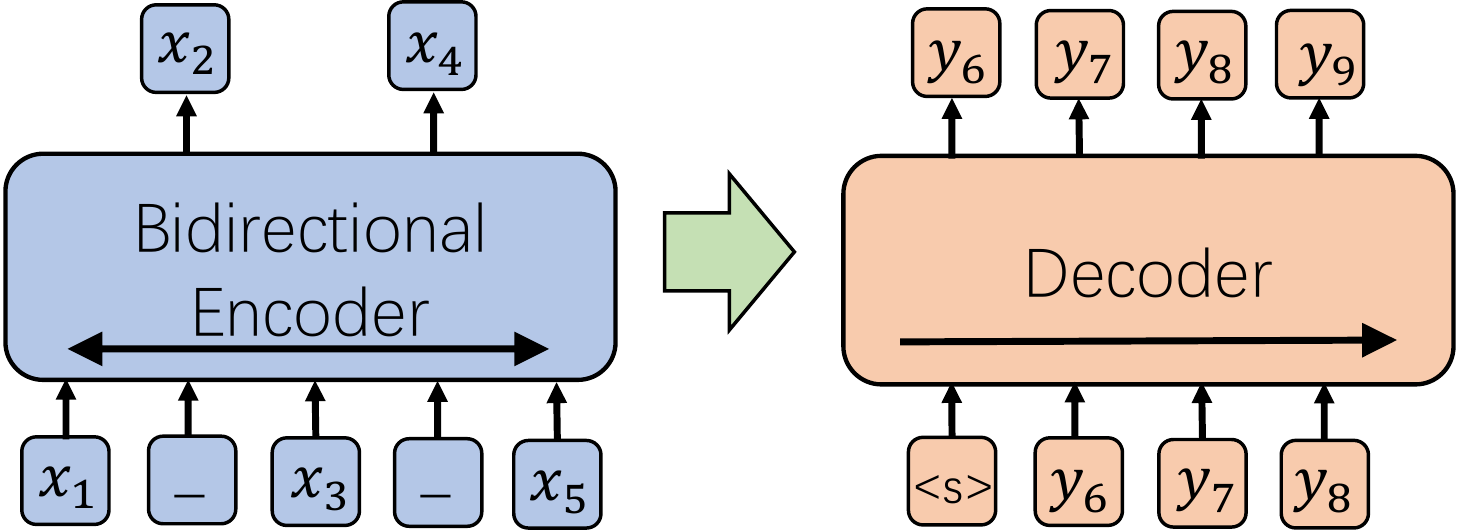}
         \caption{\method: The encoder predicts masked tokens by encoding context bidirectionally, and the decoder predicts the text segment subsequent to the context. It forces the model to learn to comprehend the context for generating relevant text.}
         \label{fig:palm}
     \end{subfigure}
     \caption{A schematic comparison of \method with GPT, MASS and BART.}
     \label{fig:comparison}
\end{figure*}

To close the gap, \method is carefully designed to autoregressively generate a text sequence by comprehending the given context in a bidirectional autoencoding manner. In particular, \method delegates autoencoding-based comprehension to the encoder in Transformer, and autoregressive generation to the Transformer decoder. The encoder and decoder are jointly pre-trained in two stages:
\begin{enumerate}
    \item The encoder is first trained as a bidirectional autoencoder to reconstruct the original text from corrupted context in which random tokens are sampled and replaced with \texttt{[MASK]} symbols following BERT's practice~\cite{bert2018jacob}. The training optimizes the cross-entropy reconstruction loss between encoder's output and original context, as Masked Language Modeling (MLM) in BERT. By predicting the actual tokens in context that are masked out, \method forces the encoder to comprehend the meaning of the unmasked tokens and the full context.
    \item The encoder and decoder are then jointly trained to autoregressively generate text output out of the context representations from the encoder. The training maximizes the log-likelihood of the text in ground truth from the decoder's output:
    \begin{equation}
        \mathcal{L}(\theta)=\sum_{(x,y)\in (\mathcal{X},\mathcal{Y})}\log \prod_{t=1}^n P(y_t|y_{<t},x;\theta),
    \end{equation}
    where $\mathcal{X}$ represents the set of context and $\mathcal{Y}$ represents the set of text to be generated. By conditioning the generation on context representations, \method forces the decoder to rely deeply on the context instead of preceding generated tokens in next token prediction, which facilitates context-sensitive generation.
\end{enumerate}

\subsection{Input\&Output Representations}
\label{sec:in-out}
In the phase of model pre-training, input and output representations are tailored to minimize the discrepancy between self-supervised pre-training and supervised fine-tuning. In a typical downstream generation task (e.g., abstractive summarization and generative QA), context is given as a rather long passage, and a model is asked to generate a shorter piece of text based on the comprehension of the context.

Given a contiguous text fragment of length $L$ (composed of a few sentences) from an unlabeled corpus, \method uses the consecutive span of length $80\%\cdot L$ from the beginning of the fragment as context input to the encoder, and uses the remainder of text span of length $20\%\cdot L$ as text output to be generated by the decoder. This representation design mimics the input and output of downstream tasks, with the hypothesis that human-written text is coherent and thus the subsequent text span of length $20\%\cdot L$ captures the comprehension of the preceding context span. In this way, \method learns to infer the subsequent text content from the preceding content.

The collection of text fragments are constructed from a corpus by following the practice of BERT. In our experiments, we set the maximum length of a fragment to be 500, i.e., $L\leq 500$. Therefore, the context input consists of at most 400 tokens, and the text output consists of at most 100 tokens.

Figure~\ref{fig:comparison} shows a schematic comparison of input\&output representations between \method and the existing pre-training generation methods, GPT, MASS and BART. GPT uses a decoder to predict tokens autoregressively, without an encoder to condition generation on context. MASS and BART are both trained to recover the original tokens that are masked out from corrupted text, where the inputs to the encoder and the decoder come from the same text segment (e.g., the sequence $(x_1,x_2,x_3,x_4,x_5)$ in Figures~\ref{fig:mass} and~\ref{fig:bart}). They are also expected to output the tokens from the same text sequence. By contrast, in \method the encoder and the decoder take two different inputs. The input to the decoder comes from the continuation of the text input to the encoder (e.g., $(y_6,y_7,y_8)$ is subsequent to $(x_1,x_2,x_3,x_4,x_5)$ in the contiguous text segment $(x_1,x_2,x_3,x_4,x_5,y_6,y_7,y_8)$ in Figure~\ref{fig:palm}). In addition to the continuation predicted by the decoder, \method produces an extra output from the encoder, which contains the predicted tokens masked in the input (e.g., $x_2$ and $x_4$ in Figure~\ref{fig:palm}). The output predictions from the encoder and the decoder are used for training in the two stages, respectively.

\subsection{Copying Tokens from Context}
In a human-written document, subsequent text often refers back to entities and tokens present earlier in the preceding text. Therefore, it would increase coherence of text generated in downstream to incorporate the copy mechanism into pre-training on an unlabeled corpus. This allows the model to learn from pre-training when and how to copy tokens in generating text, and the knowledge is transferred to downstream fine-tuning.

\begin{figure}[t]
    \centering
    \includegraphics[width=\linewidth]{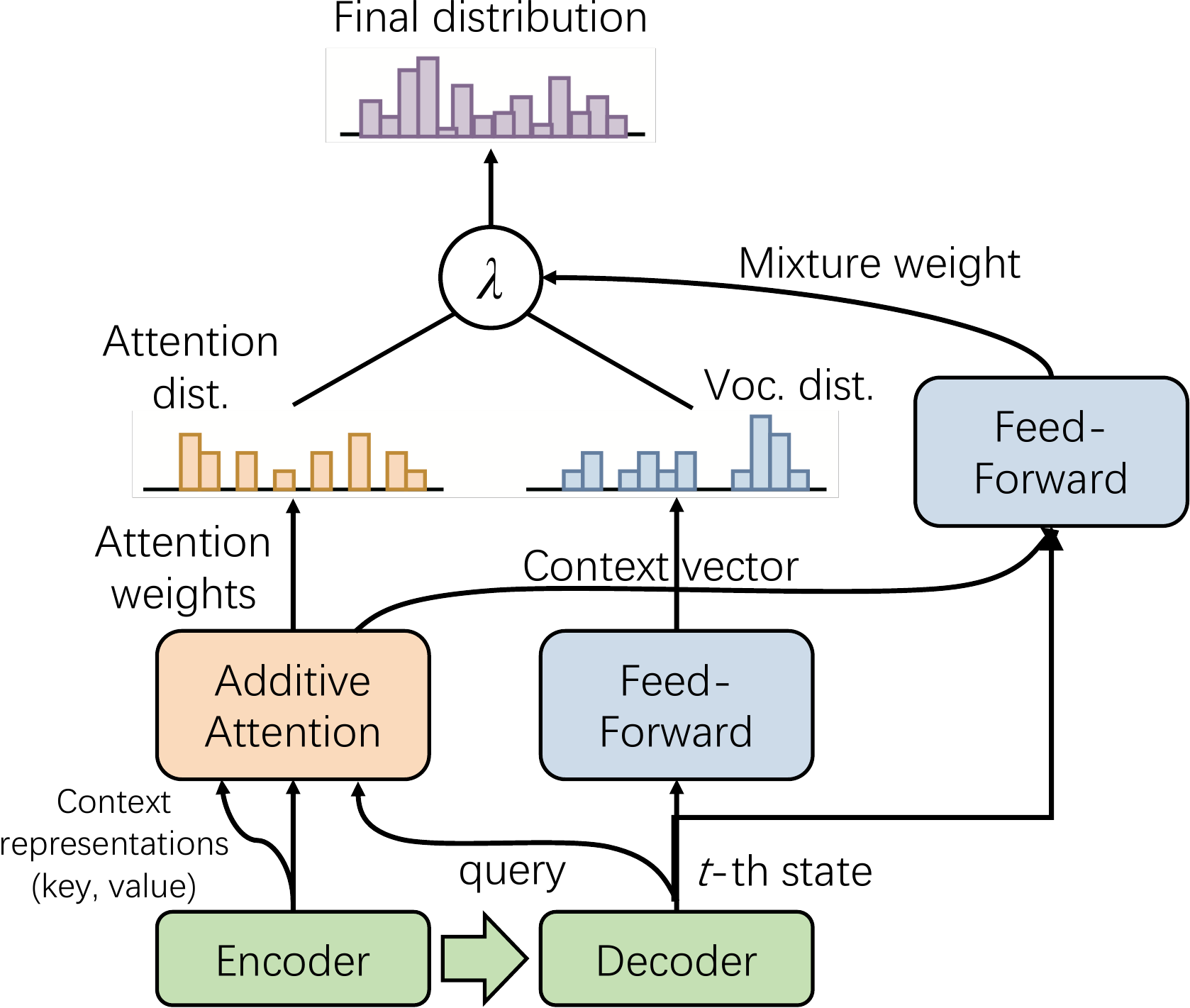}
    \caption{The pointer-generator network on top of the decoder in Transformer. For each decoding step $t$, mixture weights $\lambda$ for the probability of generating tokens from the vocabulary and copying tokens from context are calculated. The two distributions are summed in a weighted manner to obtain the final distribution.}
    \label{fig:pointer}
\end{figure}

\method incorporates the copy mechanism by plugging in the pointer-generator network~\cite{See:2017,Nishida:2019} on top of the decoder in Transformer. Figure~\ref{fig:pointer} illustrates the pointer-generator network, which allows every token to be either generated from a vocabulary or copied from context in generating text.

\textbf{Extended vocabulary distribution}. Let the extended vocabulary, $V$, be the union of words in the vocabulary and all tokens present in context. $P^v(y_t)$ then denotes the probability distribution of the $t$-th word token, $y_t$, over the extended vocabulary, defined as:
\begin{equation}
    P^v(y_t) = \text{softmax}(W^e(W^v s_t + b^v)),
\end{equation}
where $s_t$ denotes the output representation of $t$-th token from the decoder. The output embedding $W^e$ is tied with the corresponding part of the input embedding~\cite{Inan:2017}, and $W^v$ and $b^v$ are learnable parameters.

\textbf{Copy distribution}. \method uses an additional attention layer for the copy distribution on top of the decoder. In the course of generation, the layer takes $s_t$ as the query, and outputs $\alpha_t$ as the attention weights and $z_t^c$ as the context vector:
\begin{align}
e_{tl}^c&={w^c}^\top \tanh(W^m h_l^c + W^s s_t + b^c),\\
\alpha_t^c&=\text{softmax}(e_t^c),\\
z_t^c&=\sum_{l=1}^m \alpha_{tl}^c h_l^c,
\end{align}
where $h_l^c$ is the representation of $l$-th token in context from the encoder. $w^c$, $b^c$, $W^m$ and $W^s$ are learnable parameters. As a result, $P^c(y_t)$ is the copy distribution over the extended vocabulary, defined as:
\begin{equation}
    P^c(y_t)=\sum_{l:x_l=y_t} \alpha_{tl}^c.
\end{equation}

\textbf{Final distribution}. The final probability of generating $y_t$ is defined as a mixture of the extended vocabulary distribution and the copy distribution:
\begin{align}
    P(y_t)&=\lambda P^v(y_t) + (1-\lambda) P^c(y_t),\\
    \lambda&=\text{sigmoid}(w^z z_t^c + w^s s_t + b^m),
\end{align}
where $w^z$, $w^s$ and $b^m$ are learnable parameters.

The parameters in pointer-generator learned in pre-training are all kept and passed downstream for fine-tuning on labeled data.

\begin{table*}[t]
\center\small
\begin{tabular}{ c | l }
\hline\hline
\multicolumn{2}{l}{\textbf{Example 1}}\\
\hline
\multirow{5}{*}{\textbf{Input}} & A classic Aston Martin once owned by Spartacus star Peter Ustinov is set to fetch more than \pounds 1 million at auction\\
& - twice what it fetched four years ago. The actor bought the Aston Martin DB4 Cabriolet in 1962, shortly after\\
& winning a Best Supporting Actor Oscar for his role as Batiatus in Spartacus. It was one of the most luxurious cars\\
& of its day, costing \pounds 4,000 and was delivered to him at a Swiss hotel at a time when the average house price in\\
& Britain was just \pounds 2,500.\\
\hline
\multirow{2}{*}{\textbf{\method}} & The Aston Martin DB4 Cabriolet was bought by Peter Ustinov for \pounds 4000 and was expected to fetch for \pounds 2.5\\
& million for auction. The car was sold for \pounds 1.2 million finally.\\
\hline
\textbf{MASS} & peter ustinov's \texttt{UNK} auctioned for more than \$1 million.\\
\hline\hline
\multicolumn{2}{l}{\textbf{Example 2}}\\
\hline
\multirow{6}{*}{\textbf{Input}} & Cape Verde's 2-0 win over Portugal was the most eye-catching international result of the week. So, who are Cape\\
& Verde and why has this tiny island off the west coast of Africa suddenly become an international football force?\\
& Where are the Cape Verde Islands? Cape Verde is a group of islands 400 miles from Senegal off the west coast of\\
& Africa. Its population is around 500,000 and boasts both beautiful beaches and striking volcanic landscapes,\\
& making it a haven for tourism.\\
\hline
\multirow{3}{*}{\textbf{\method}} & Cape Verde is a small island off the west coast of Africa with a population of around 500,000 and boasts both\\
& beautiful beaches and striking volcanic landscapes, making it a haven for tourism. Cape Verde is home to the\\
& Cape Verde Islands National Park with a number of islands.\\
\hline
\textbf{MASS} & tiny african island nation cape verde has beautiful beaches.\\
\hline\hline
\end{tabular}
\caption{Example generated continuations of the text input to \method and MASS.}
\label{table:pretraining}
\vspace{-10pt}
\end{table*}

\section{Experiments}
In this section, we present the experimental setup and results of \method pre-training on a large unlabeled corpus and fine-tuning on a variety of language generation tasks, including generative QA, abstractive summarization, question generation, and conversational response generation.


\subsection{Pre-training Configuration}
\textbf{Experimental Setup}. \method is based on the Transformer which consists of a 12-layer encoder and a 12-layer decoder with 768 embedding/hidden size, 3072 feed-forward filter size and 12 attention heads. We have also trained a larger model, referred to as \method$_{\text{LARGE}}$, to compare with the baseline models of the same size. \method$_{\text{LARGE}}$ has an encoder of 24 layers and a decoder of 6 layers, with 1024 embedding/hidden size and 16 attention heads. The parameters of PALM's encoder are initialized by the pre-trained RoBERTa model\footnote{\url{https://github.com/pytorch/fairseq}} which was trained with the Masked LM objective, removing Next Sentence Prediction from BERT.

\method is trained with a dropout rate of 0.1 on all layers and attention weights, and a GELU activation function~\cite{Hendrycks:2016} used as GPT. The learning rate is set to 1e-5, with linear warmup over the first 10k steps and linear decay. The pre-training procedure runs on 16 NVIDIA V100 GPU cards for 800K steps, with each minibatch containing 64 sequences of maximum length 500 tokens.

\textbf{Pre-training Dataset}. We use documents of English Wikipedia and BookCorpus~\cite{DBLP:journals/corr/ZhuKZSUTF15} as our pre-training corpus, and perform WordPiece tokenization as BERT~\cite{bert2018jacob}. The documents are split into sentences. Different from BERT, we use multiple consecutive sentences up to 400 tokens as the source text input to the encoder, and use the subsequent consecutive sentences up to 100 tokens as the target text to the decoder. The pre-training dataset $(\mathcal{X},\mathcal{Y})$ is constructed from the documents by a sliding window with the stride of one sentence, resulting in 50M $(x,y)$ pre-training pairs.

\subsection{Unsupervised Pre-training}
To understand the performance of \method pre-training, we compare generation quality of the pre-trained models of \method and MASS~\footnote{\url{https://modelrelease.blob.core.windows.net/mass/mass_summarization_1024.pth}}. Specifically, we feed a few sentences from a news article to both pre-trained models, and the models generate a continuation of the input sentences by beam search with a beam of size 5. The news articles from CNN~\footnote{\url{https://drive.google.com/uc?export=download&id=0BwmD_VLjROrfTHk4NFg2SndKcjQ}} are used as input text to eliminate the possibility of the text present in the models' pre-training corpus, i.e., Wikipedia and BookCorpus.

The overall perplexity of \method is 17.22, which is much better than MASS's perplexity of 170.32, indicating \method's better language modeling. Table~\ref{table:pretraining} illustrates a couple of example continuations generated by \method and MASS. In both examples, \method generates fluent and grammatical English, while MASS outputs a short sentence that is much less relevant to input text, since the MASS model was trained on individual sentences. In the first example, it is interesting to observe that in addition to summarizing the input content, \method is able to make a non-trivial inference of the expected auction price and the final selling price of the car (might not be factually accurate though). An inference is also made by \method in the second example in addition to summarization, although the Cape Verde Islands National Park does not really exist.

These examples demonstrate that \method pre-training has learned to infer and to reason from the input text. Although in the pre-training phase the generated content may not be factually accurate in the absence of rich context, the capability of inference can be transferred downstream by fine-tuning on specific generation tasks.

\subsection{Fine-tuning on Generative QA}
We also experiment with fine-tuning \method on several downstream generation tasks. The MARCO benchmark~\cite{Nguyen:2016} released by Microsoft is a good fit for evaluating generative QA models. In the MARCO dataset, the questions are user queries issued to the Bing search engine and the contextual passages are from real web documents. The data has been split into a training set (153,725 QA pairs), a dev set (12,467 QA pairs) and a test set (101,092 questions with unpublished answers). To evaluate the generative capability, we focus on the \emph{Q\&A + Natural Language Generation} task, the goal of which is to provide the best answer available in natural language that could be used by a smart device / digital assistant.

The answers are human-generated and not necessarily sub-spans of the contextual passages, so we use the ROUGE-L~\cite{Lin:2004} metric for our evaluation to measure the quality of generated answers against the ground truth.

We fine-tune the pre-trained \method on the MARCO training set for 10 epochs. We set the batch size to 64, the learning rate to 1e-5, and the maximum input length to 512. The other hyper-parameters are kept the same as pre-training. In fine-tuning \method, the encoder takes as input $x$ a contextual passage concatenated with a question at the end, and the decoder takes an answer as input $y$. During decoding, we use beam search with a beam of size 5.

\begin{table}[t]
\center
\begin{tabular}{ l | c}
\hline\hline
\textbf{Method} & \textbf{Rouge-L}\\
\hline\hline
ConZNet~\cite{Indurthi:2018} & 0.421\\
\hline
Reader-Writer & 0.439\\
\hline
KIGN-QA & 0.441\\
\hline
SNET+CES2S & 0.450\\
\hline
Communicating BERT & 0.483\\
\hline
VNET~\cite{VNET2018} & 0.484\\
\hline
Selector NLGEN & 0.487\\
\hline
BERT+Multi-Pointer & 0.495\\
\hline
Masque~\cite{Nishida:2019} & 0.496\\
\hline
\textbf{\method} & \textbf{0.498}\\
\hline\hline
\end{tabular}
\caption{Test results of answer generation on the official MARCO leaderboard as of December 9, 2019.}
\label{table:marco}
\vspace{-10pt}
\end{table}

\begin{table*}[t]
\center
\begin{tabular}{ l c c c c c c}
\hline\hline
\textbf{} & \multicolumn{3}{c}{CNN/DailyMail} &  \multicolumn{3}{c}{Gigaword} \\

\textbf{} & \textbf{RG-1} & \textbf{RG-2} & \textbf{RG-L} 
& \textbf{RG-1} & \textbf{RG-2} & \textbf{RG-L}\\
\hline
BERTSUMABS \cite{liu-lapata-2019-text} & 41.72 & 19.39 & 38.76 & - & - & - \\

MASS \cite{mass2019song} & 42.12 & 19.50 & 39.01 & 38.13 & 19.81 & 35.62 \\

UniLM$_{\text{LARGE}}$ \cite{unilm2019} & 43.33 & 20.21 & 40.51 & 38.45 & 19.45 & 35.75  \\

T5$_{\text{LARGE}}$ \cite{raffel2019exploring} & 42.50 & 20.68 & 39.75 & - & - & - \\

BART$_{\text{LARGE}}$ \cite{bart2019} & 44.16 & 21.28 & 40.90 & - & - & - \\

PEGASUS \cite{pegasus} & 44.17 & \textbf{21.47} & 41.11 & 39.12 & 19.86 & 36.24  \\

ERNIE-GEN$_{\text{LARGE}}$ \cite{ernie-gen} & 44.02 & 21.17 & 41.26 & 39.25 & 20.25 & 36.53 \\

\hline

\textbf{\method} & 42.71 & 19.97 & 39.71 & 38.75 & 19.79 & 35.98 \\
\textbf{\method}$_{\text{LARGE}}$ & \textbf{44.30} & 21.12 & \textbf{41.41} & \textbf{39.45} & \textbf{20.37} & \textbf{36.75} \\
\hline\hline
\end{tabular}
\caption{Results of abstractive summarization on the CNN/DailyMail test set and the Gigaword test set. RG is short for ROUGE}
\label{table:gigaword}
\vspace{-10pt}
\end{table*}

\begin{table}[b!]
\vspace{-5pt}
\center
\begin{tabular}{ l | c | c | c}
\hline\hline
{\textbf{Method}} & \textbf{BLEU-4} & \textbf{MTR} & \textbf{RG-L} \\
\hline\hline
CorefNQG$^a$ & 15.16  & 19.12 & - \\
MP-GSN$^b$ & 16.38 & 20.25 & 44.48 \\
UNILM$^c$ & 22.88 & 24.94 & 51.80 \\
ERNIE $^d$ & 22.28 & 25.13 & 50.58  \\
ERNIE-GEN$_{\text{LARGE}}$ $^d$ & 24.03 & \textbf{26.31} & 52.36 \\
\hline
\textbf{\method} & 22.78 & 25.02 & 50.96 \\
\textbf{\method}$_{\text{LARGE}}$ & \textbf{24.11} & 25.85 & \textbf{52.38} \\
\hline\hline
\end{tabular}
\caption{Question generation results on the SQuAD
dataset. MTR is short for METEOR and RG is short for ROUGE. $^a$~\cite{corefnqg}; $^b$~\cite{zhao-etal-2018-paragraph}; $^c$~\cite{unilm2019}; $^d$~\cite{ernie-gen}.}
\label{table:squad}
\end{table}

Table~\ref{table:marco} presents the answer generation results on the test set obtained from the official MARCO leaderboard. \method achieves the 1st place on the leaderboard, outperforming all competing methods in generation quality. Note that \method pre-trains a single model, while some of the top-performing methods are ensemble models, such as Masque, on the leaderboard. Crucially, the superiority of \method-single over Masque-ensemble with pre-trained ELMo~\cite{peters-etal-2018-deep} and BERT-based methods clearly demonstrates the effectiveness and generalizability of \method over the other pre-training approaches in language modeling.

\subsection{Fine-tuning on Summarization}
Text summarization produces a concise and fluent summary conveying the key information in the input (e.g., a news article). We focus on abstractive summarization, a generation task where the summary is not constrained to reusing the phrases or sentences in the input text. We conduct experiments on both the CNN/DailyMail dataset~\cite{karl2015teaching} and the Gigaword dataset~\cite{gigaword}. The CNN/DailyMail dataset contains 93K news articles from CNN and 220K articles from Daily Mail, while the Gigaword dataset consists of a total of 3.8M article-title pairs. We take the articles as the input to the encoder and the summary for the decoder. We adopt the same optimization hyperparameters from generative QA fine-tuning for the summarization task. The F1 scores of Rouge-1, Rouge-2 and Rouge-L are reported on the test set of both datasets for evaluation.

Table~\ref{table:gigaword} shows the results of abstractive summarization on the CNN/DailyMail test set and the Gigaword test set. PALM achieves better performance than all strong summarization models with pre-training recently proposed, including UniLM~\cite{unilm2019}, T5~\cite{raffel2019exploring}, BART~\cite{bart2019}, PEGASUS~\cite{pegasus} and ERNIE-GEN~\cite{ernie-gen}. By consistently outperforming the pre-training methods, \method confirms its effectiveness in leveraging unsupervision signals for language generation.

\subsection{Fine-tuning on Question Generation}
We conduct experiments for the answer-aware question generation task. Given an input passage and an answer span, question generation aims to generate a question that leads to the answer. Following the practice in~\cite{zhao-etal-2018-paragraph,unilm2019}, we use the SQuAD 1.1~\cite{squad} dataset, and the BLEU-4, METEOR and ROUGE-L metrics for evaluation.

As shown in Table~\ref{table:squad}, \method outperforms all previous question generation systems and achieves a new state-of-the-art result on BLEU-4 and ROUGE-L for question generation on the SQuAD 1.1 dataset.

\subsection{Fine-tuning on Response Generation}
\vspace{-5pt}
Conversational response generation aims to produce a flexible response to a conversation~\cite{Vinyals:2015}. Following MASS, we conduct experiments on the Cornell Movie Dialog corpus\footnote{\url{https://github.com/suriyadeepan/datasets/tree/master/seq2seq/cornell_movie_corpus}}~\cite{Danescu:2011} that contains 140K conversation pairs, and use the training/test splits provided by the dataset. The same training hyperparameters from generative QA fine-tuning are adopted on the response generation task. We report the results in perplexity following~\cite{Vinyals:2015} (lower is better).

We compare \method with the competing methods including the baseline trained on the data pairs available and the pre-trained BERT+LM and MASS. Following MASS, we train every model on 10K pairs randomly sampled and all 110K training pairs. As shown in Table~\ref{table:response}, \method significantly performs better than all the competitors by a large margin on both the 10K and 110K data, demonstrating its capability in generating responses to context thanks to its new pre-training objectives.

\vspace{-5pt}
\subsection{Ablation Studies}
\vspace{-5pt}
We conduct ablation studies to assess the individual contribution of every component in \method. Table~\ref{table:ablation} reports the results of full \method and its ablations on the CNN/Daily Mail summarization dataset.

We evaluate how much the pointer-generator network contributes to generation quality by removing it from \method pre-training. This ablation results in a drop from 39.71 to 39.49 on Rouge-L, demonstrating the role of the pointer-generator in generative modeling. Given the slight drop, one may choose to exclude it from the full model for training efficiency. In our experiments, the pointer-generator is used in every generation task for optimal generation performance.

To study the effect of the pre-trained encoder and decoder in \method, we ablate autoencoding and autoregression by randomly initializing the weights of the encoder and the decoder, respectively. The autoencoding and autoregression components both prove to be critical with significant drops on the three Rouge metrics after the ablation. Finally, we study the significance of full \method pre-training. Over 6.5\% of performance degradation resulted from ablating pre-training clearly demonstrates the power of \method in leveraging an unlabeled corpus for downstream generation.

\begin{table}[t]
\center
\begin{tabular}{ l | c | c }
\hline\hline
\multirow{2}{*}{\textbf{Method}} & \textbf{Perplexity} & \textbf{Perplexity}\\
& \textbf{(10K Data)} & \textbf{(110K Data)}\\
\hline\hline
Baseline & 82.39 & 26.38\\
\hline
BERT+LM & 80.11 & 24.84\\
\hline
MASS & 74.32 & 23.52\\
\hline
\textbf{\method} & \textbf{45.43} & \textbf{21.98}\\
\hline\hline
\end{tabular}
\caption{Results of conversational response generation in terms of perplexity on Cornell Movie Dialog corpus (lower is better).}
\label{table:response}
\vspace{-10pt}
\end{table}

\vspace{-5pt}
\section{Related Work}
\vspace{-5pt}
ELMo~\cite{peters-etal-2018-deep} is an early prominent pre-training method based on bidirectional LSTMs. It concatenates left-only and right-only representations, but does not pre-train interactions between these features. GPT~\cite{Radford2018ImprovingLU}, GPT-2~\cite{gpt2} and GPT-3~\cite{gpt3} are proposed to base language modeling on the Transformer architecture, and use only the Transformer decoder for pre-training. Edunov \emph{et al.}~\cite{Edunov:2019} examine different strategies (e.g., ELMo) to add contextualized embeddings to sequence-to-sequence models, and observe the most improvement by adding the learned embeddings to the encoder.

BERT~\cite{bert2018jacob} introduces Masked Language Modelling, which allows pre-training to learn interactions between left and right context words. Recent work has shown that very strong performance can be achieved by training for longer~\cite{roberta2019}, by tying parameters across layers~\cite{albert2019}, and by masking spans instead of words~\cite{spanbert2019}. However, BERT does not make predictions autoregressively, so it is not effective for generation tasks.

UniLMs~\cite{unilm2019,unilmv2} fine-tune BERT with an ensemble of masks, some of which use only leftward context, allowing UniLMs to be used for generation tasks. A difference between UniLMs and \method is that UniLMs are not fully autoregressive in the pre-training process. In contrast, \method reduces the mismatch between pre-training and context-conditioned generation tasks by forcing the decoder to predict the continuation of text input on an unlabeled corpus.

MASS~\cite{mass2019song} and BART~\cite{bart2019} are the two pre-training methods most similar to \method. In MASS, an input sequence with a masked span of tokens is mapped to a sequence consisting of the missing tokens, whereas BART is trained to reconstruct the original text from corrupted input with some masked tokens. The difference in input \& output representations between \method and MASS \& BART is detailed in Section~\ref{sec:in-out}.

\begin{table}[t]
\center
\begin{tabular}{ l | c | c | c}
\hline\hline
{\textbf{Ablation}} & \textbf{RG-1} & \textbf{RG-2} & \textbf{RG-L}\\
\hline\hline
\textbf{\method} & \textbf{42.71} & \textbf{19.97} & \textbf{39.71} \\
\hline
\xmark\ pointer-generator & 42.54 & 19.86 & 39.49 \\
\hline
\xmark\ autoencoding & 41.78 & 19.32 & 38.81 \\
\hline
\xmark\ autoregression & 41.89 & 19.48 & 38.92 \\
\hline
\xmark\ pre-training & 40.32 & 17.78 & 37.12 \\
\hline\hline
\end{tabular}
\caption{Ablation tests of \method on the CNN/Daily Mail summarization dataset.}
\label{table:ablation}
\vspace{-8pt}
\end{table}

\section{Conclusions}
\vspace{-5pt}
In this work, we propose \method, a novel approach to pre-training an autoencoding and autoregressive language model on a large unlabeled corpus, designed to be fine-tuned on downstream generation conditioned on context. It is built upon an extension of the Transformer encoder-decoder, and jointly pre-trains the encoder and the decoder in an autoencoding denoising stage followed by an autoregressive generation stage.

\method significantly advances the state-of-the-art results on a variety of context-conditioned generation applications, including generative QA (Rank 1 on the MARCO leaderboard), abstractive summarization, question generation, and conversational response generation. It has been shown in prior work~\cite{roberta2019} that training for more steps over a larger corpus can potentially improve the performance of pre-training. Our future work will explore the potential of training \method for longer on much more unlabeled text data.

\bibliography{palm}
\bibliographystyle{acl_natbib}

\end{document}